\documentclass{bmvc2k}

\usepackage{tikz}
\usepackage{comment}
\usepackage{amsmath,amssymb} 
\usepackage{color}
\usepackage{booktabs} 
\usepackage{microtype} 
\usepackage{float} 
\usepackage{paralist}
\usepackage{wrapfig}
\usepackage{capt-of}
\newcommand{\noop}[1]{}

\renewenvironment{itemize}[1]{\begin{compactitem}#1}{\end{compactitem}}
\renewenvironment{enumerate}[1]{\begin{compactenum}#1}{\end{compactenum}}

\graphicspath{{figures/}{pictures/}{images/}{./}} 

\title{CASAPose: Class-Adaptive and Semantic-\\Aware Multi-Object Pose Estimation}

\addauthor{Niklas Gard}{niklas.gard@hhi.fraunhofer.de}{1}
\addauthor{Anna Hilsmann}{anna.hilsmann@hhi.fraunhofer.de}{1}
\addauthor{Peter Eisert}{peter.eisert@hhi.fraunhofer.de}{1,2}

\addinstitution{
 Fraunhofer HHI\\
 Einsteinufer 37,\\
 Berlin, Germany
}
\addinstitution{
 Institute for Computer Science\\
 Humboldt University of Berlin,\\
 Berlin, Germany
}

\runninghead{Gard, Hilsmann, Eisert}{CASAPose}

\def\etal{\emph{et al}\bmvaOneDot}

\begin{document}

\maketitle

\begin{abstract}
Applications in the field of augmented reality or robotics often require joint localisation and 6D pose estimation of multiple objects.  
However, most algorithms need one network per object class to be trained in order to provide the best results. 
Analysing all visible objects demands multiple inferences, which is memory and time-consuming.
We present a new single-stage architecture called CASAPose that determines 2D-3D correspondences for pose estimation of multiple different objects in RGB images in one pass.
It is fast and memory efficient, and achieves high accuracy for multiple objects by exploiting the output of a semantic segmentation decoder as control input to a keypoint recognition decoder via local class-adaptive normalisation. 
Our new differentiable regression of keypoint locations significantly contributes to a faster closing of the domain gap between real test and synthetic training data.
We apply segmentation-aware convolutions and upsampling operations to increase the focus inside the object mask and to reduce mutual interference of occluding objects.
For each inserted object, the network grows by only one output segmentation map and a negligible number of parameters. 
We outperform state-of-the-art approaches in challenging multi-object scenes with inter-object occlusion and synthetic training.
Code is available at: \url{https://github.com/fraunhoferhhi/casapose}.
\end{abstract}

\section{Introduction}
\begin{figure}[t]
\centering
  \includegraphics[width=.96\columnwidth]{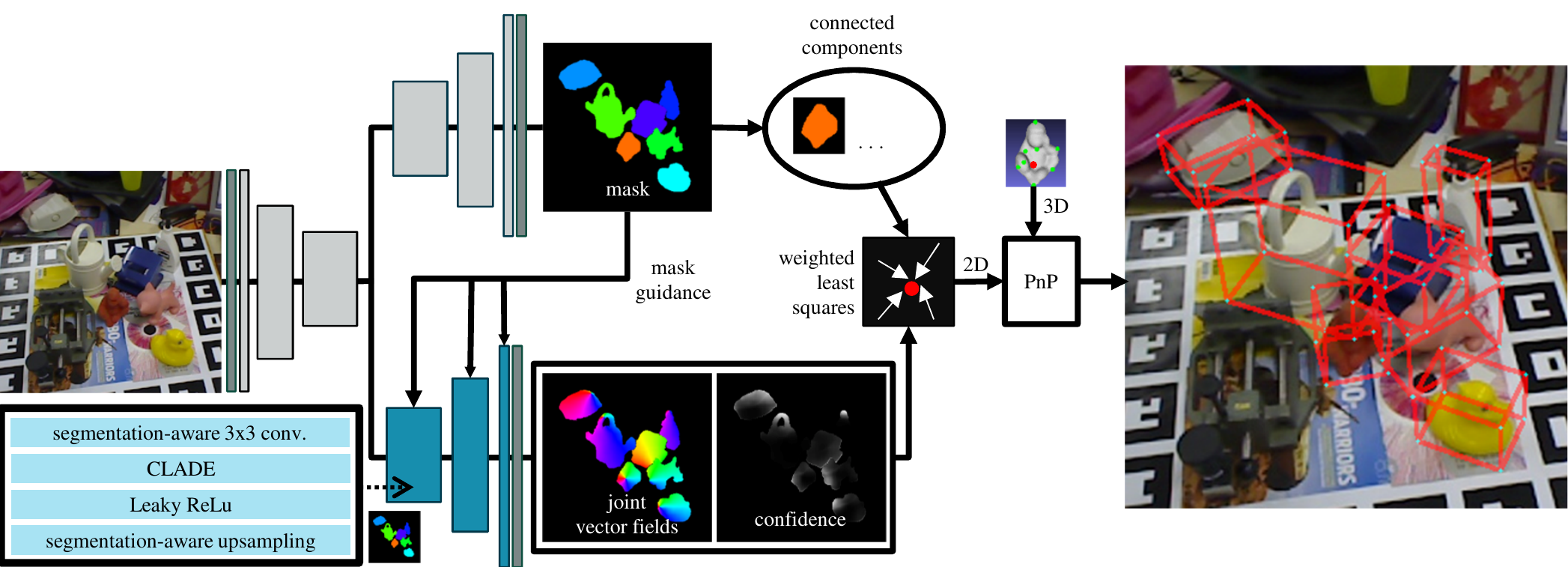}
 \caption{In CASAPose, a segmentation decoder estimates object masks that guide a second decoder to predict 2D-3D correspondences for single-stage multi-object pose estimation.}
 \label{fig:strcuture}
\end{figure}
In this paper, we present CASAPose, a novel architecture specifically designed to improve multi-object scalability of a 6D pose estimator. Retrieving the pose of objects in front of a camera in real-time is essential for augmented reality and robotic object manipulation. Convolutional neural networks (CNN) led to a significant boost in accuracy and robustness against occlusion or varying illumination. Many methods train a separate CNN per object \cite{Peng2019,Song2020,Tremblay2018,Park2019_2,Zhigang2019,Sundermeyer2019}. In scenarios with multiple objects, this leads to impractical side effects. If it is not known which objects are present in the image, it must either be inspected individually for each known object or an additional detector is required for identification. Performing inference for each visible object increases the computation time. In addition, the need for multiple networks increases memory usage and lengthens training. Pose estimation is often formulated as an extension of semantic segmentation, with additional output maps to infer the pose from, e.g.~3D object coordinates, or keypoint locations encoded in vector fields. The trivial multi-object extension to expand this secondary output for each object \cite{Peng2019} results in GPU-intensive slow training and performance degradation \cite{Gard2022}.

Our approach decouples semantic object identification and keypoint regression and uses the pixel-wise segmentation results as guidance for a keypoint predicting branch. We introduce techniques from GAN-based conditional image synthesis and style transfer to the field of object pose estimation. First, we improve the descriptive capability of the network by adding a small amount of class specific extra weights to the network. Applying these parameters locally in the decoder as a class-adaptive de-(normalisation) (CLADE) \cite{Tan2020} adds semantic awareness to the network. The locations of 3D keypoints on the object, represented by vectors pointing towards their 2D projection are interpreted as a local object-specific style. Second, we strengthen the local focus of the network by integrating two more semantic-aware operations \cite{Dundar2020}. A guided convolution re-weights the convolutions in the keypoint branch and forces the decoder to focus on the mask region, when estimating the 6D pose. A segmentation-aware upsampling uses the mask during upscaling of feature maps to avoid misalignments between low-resolution features and high-resolution segmentation maps. Both operations reduce the interference between mutually occluding objects and improve the quality of the pose estimates. From the clearly separated output, the 2D-3D correspondences are localised in a robust and differentiable manner via weighted least squares.
Since dataset creation for multi-object scenarios is very time consuming, we train only on synthetic data that comes with free labels and can be generated with free tools. To summarise, we make the following contributions:

\begin{enumerate}  
\item We show that incorporating a small set of object-specific parameters through CLADE significantly increases the multi-object capacity of a pose estimation CNN. 
\item We reduce the number of outputs of encoder-decoder based pose estimation networks and reduce intra-class interferences by segmen\-tation-aware transformations.
\item We exploit the strictly local feature processing to obtain a new differentiable 2D keypoint estimation method improving the accuracy of 2D-3D correspondences.
\item We train our network only on synthetic images and achieve state-of-the art results on real data.
\end{enumerate}

\section{Related Work}
\label{sec:related work}
\paragraph{Multi-Object 6D Pose Estimation}
Approaches for 6D pose estimation with CNNs usually either regresses the object pose directly \cite{Xiang2018,Billings2019,Wang2020}, describe the object's appearance with a latent space code to compare it to pre-generated codes \cite{Sundermeyer2019,Wen2020, Park2020}, or regresses the position of 2D projections of 3D points and calculate the pose with a Perspective-n-Points (PnP) algorithm. Approaches from last category either predict object specific keypoints \cite{Hu2019,Peng2019,Song2020,Tremblay2018,Rad2017} or dense correspondence/coordinate maps \cite{Zakharov2020,Park2019_2,Hou2020,Thalhammer2021,Zhigang2019,Hodan2020}.

To deal with multiple objects, most approaches train a separate network per object and need multiple inferences per image \cite{Peng2019,Song2020,Tremblay2018,Park2019_2,Zhigang2019,Sundermeyer2019}. Alternatively, increasing the number of output maps is proposed as multi-object extension \cite{Peng2019,Rad2017,Hodan2020}, which risks serious accuracy drops \cite{Sock2020} or complex and slow processing \cite{Hodan2020}. Each added object contributes multiple extra output channels having the same size as the input image, which requires significant GPU memory and complicates training \cite{Gard2022}. Sock \etal \cite{Sock2020} add additional weights to optimise a CNN for multiple objects and reduce the multi-object performance gap. Still, they need the object class as input from a separate bounding box detector to perform their re-parametrisation and one inference for every visible object. This principle is also applied by other multi-stage methods using object specific networks per detection \cite{Zhigang2019,Billings2019,Park2019_2, Sundermeyer2019, Zhang2021, Yang2021}. 

Other single-stage - multi-object strategies have rarely been discussed. The category level approach by Hou \etal \cite{Hou2020} unifies features from different instances of one class. It requires approximately similar geometric structure per category and aligned models during training. 
Similar to us, two recent works \cite{Thalhammer2021, Aing2021} make single-stage multi-object pose estimation more performant by using a patch-based approach on a specialised feature pyramid network \cite{Thalhammer2021}, or by predicting an error mask to filter faulty pixels near silhouettes \cite{Aing2021}. Especially for multi-object approaches, the use of synthetic training data is of high importance due difficult creation of real datasets \cite{Thalhammer2021, Hodan2020}. It offers difficulties due to the domain gap \cite{Wang2020, Yang2021, Zhang2021}, which we narrow down by giving the network access to the silhouettes, a nearly domain invariant feature \cite{Wen2020, Billings2019}.
\paragraph{Conditional Normalisation} \quad
Normalisation layers in CNNs speed-up the training and improve the accuracy \cite{Ioffe2015}. A learnable affine transformation recentres and rescales the normalised features. In the unconditional case, the normalisation does not depend on external data \cite{Ioffe2015, Ulyanov2016}. Conditional Instance Normalisation (CIN)  \cite{Dumoulin2017} increases the capacity of a CNN by learning multiple sets of normalisation parameters for different classes, e.g.~for neural style transfer \cite{Gatys2015}. 
Sock \etal apply CIN on multi-object 6D pose estimation \cite{Sock2020}, but require object identity as input and handle only one identity at a time. Spatially-adaptive instance (de)norma\-lisation (SPADE) \cite{Park2019} uses per-pixel normalisation parameters depending on semantic segmentation and a pixel’s position.
Tan \etal \cite{Tan2020} reduce computational and parameter overhead of SPADE by prioritising semantic over spatial awareness. In their class-adaptive instance (de)norma\-lisation (CLADE), a guided sampling operation selects the set of de-normalisation parameters based on the semantic class of a pixel. Similar to \cite{Gard2022}, we extend \cite{Sock2020} by first estimating a semantic segmentation and then infusing the CIN parameters on a local per-pixel level with a CLADE layer in a single pass. 

\paragraph{Content-aware Transformations}
While the spatial invariance of convolutions is beneficial for most computer vision tasks, sometimes local awareness of a filter can be helpful. CoordConv \cite{Liu2018} demonstrates the benefit of giving a filter access to its position. It has been used for panoptic segmentation \cite{Sofiiuk2019,Wang2019} and semantic image synthesis \cite{Tan2020}. Other approaches use binary maps to mask out regions of the feature map, e.g.~for inpainting, depth upsampling, or padding \cite{Liu2018_3,Uhrig2017,Liu2018_2}. Guided convolutions have been extended for non-binary annotations \cite{Yu2019} and adapted for multi-class image synthesis \cite{Dundar2020}.
Mazzini \etal \cite{Mazzini2018} point out that spatially invariant operations for feature-map upsampling fail to capture the semantic information required by dense prediction tasks. Guided operations make upsampling learnable \cite{Wang2019_2,Mazzini2018}. The content-aware upscaling by Dundar \etal \cite{Dundar2020} takes advantage of a higher resolution mask to keep features aligned with the instance segmentation. 
We apply guided convolution and upsampling \cite{Dundar2020} in our segmentation-aware decoder to force the network to focus on the object region when inferring the keypoint locations. This strengthens the local influence of the CLADE parameters inside the respective segmented region.

\section{Multi-Object Pose Estimation}
\label{sec:method}
CASAPose is an encoder-decoder structure with two decoders applied successively (Fig.~\ref{fig:strcuture}). 
The first estimates a segmentation mask that guides the second in estimating vectors pointing to 2D projections of predefined 3D object keypoints \cite{Peng2019}. For each pixel, the set of keypoints corresponds to its semantic label. The second decoder additionally outputs confidences from which we calculate the common intersection points with weighted least squares (LS). The differentiable operation allows direct optimisation of the intersection point in the loss function. 

\subsection{Object-adaptive Local Weights}
\label{sec:clade}
The injection of object-specific weights is intended to enhance the capacity of a pose estimation network for multiple objects. The process is therefore divided into two subtasks. First, a semantic segmentation $m \in \mathbb{L}^{H\times W}$ is estimated to specify each pixel's object class, out of a set $\mathbb{L}$ with ${N_c}$ class indices, in an image of size $H\times W$. Second, a semantic image synthesis, conditioned on the semantic segmentation, generates a set of vector fields specifying 2D-3D correspondences of object keypoints to calculate poses from.

 After a convolution, a conditional normalisation layer normalises each channel $k$ out of $N_k$ channels of features $x$ with its mean $\mu_k$ and standard deviation $\sigma_k$. The result is modulated with a learned scale $\gamma_k$ and shift $\beta_k$ depending on a condition $l = 1, \dots , N_c$.

\begin{equation}
x^{out}_k = \gamma^{l}_k  \frac{x_k - \mu_k}{\sigma_k} + \beta^{l}_k
\end{equation}

\begin{figure}[t!]
\centering     
\subfigure[CLADE using Guided Sampling.]{\label{fig:clade}\includegraphics[height=.352\linewidth]{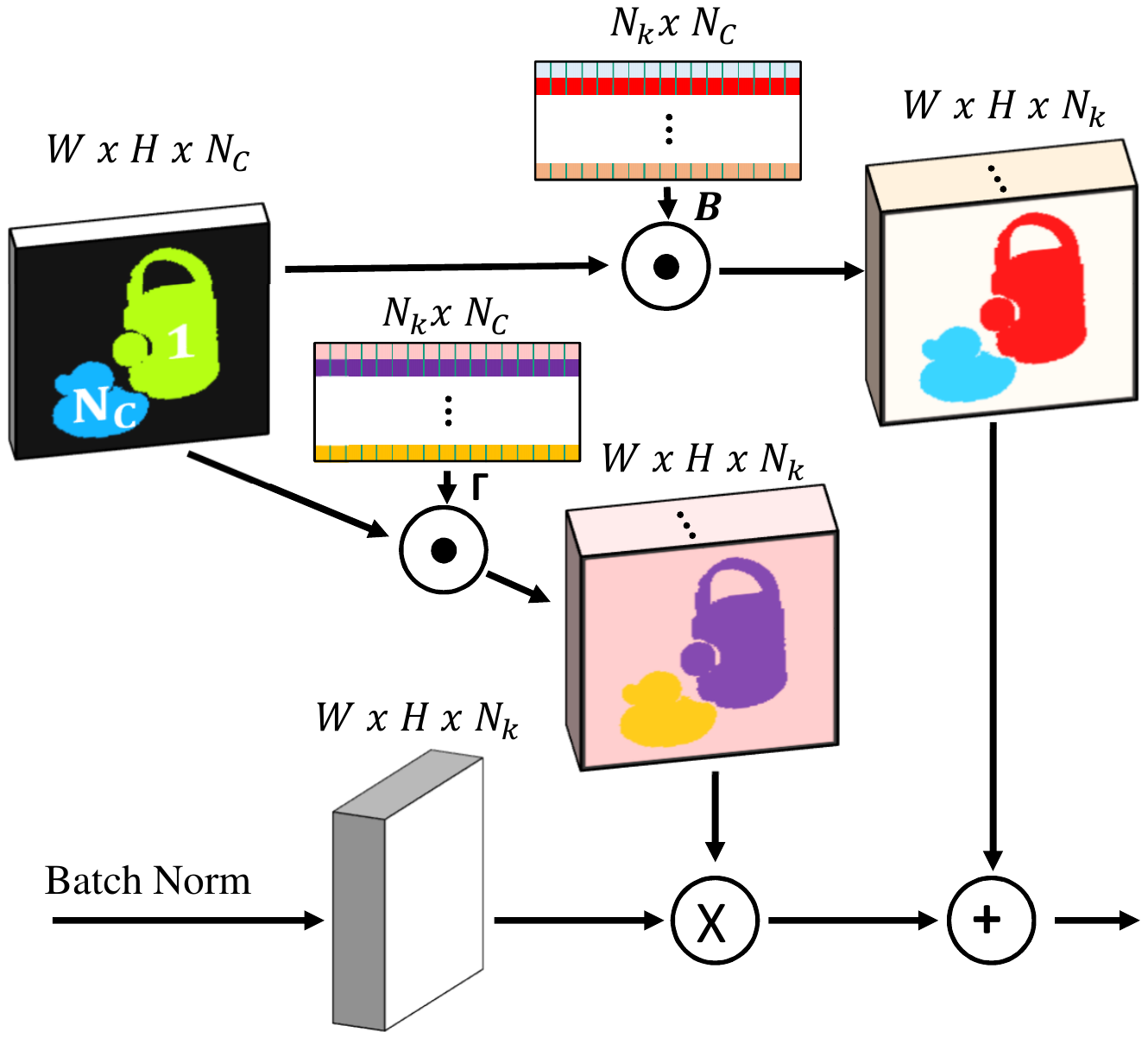}}
\quad \quad \quad
\subfigure[Guided convolution and upsampling.]{\label{fig:guided}\includegraphics[height=.33\linewidth]{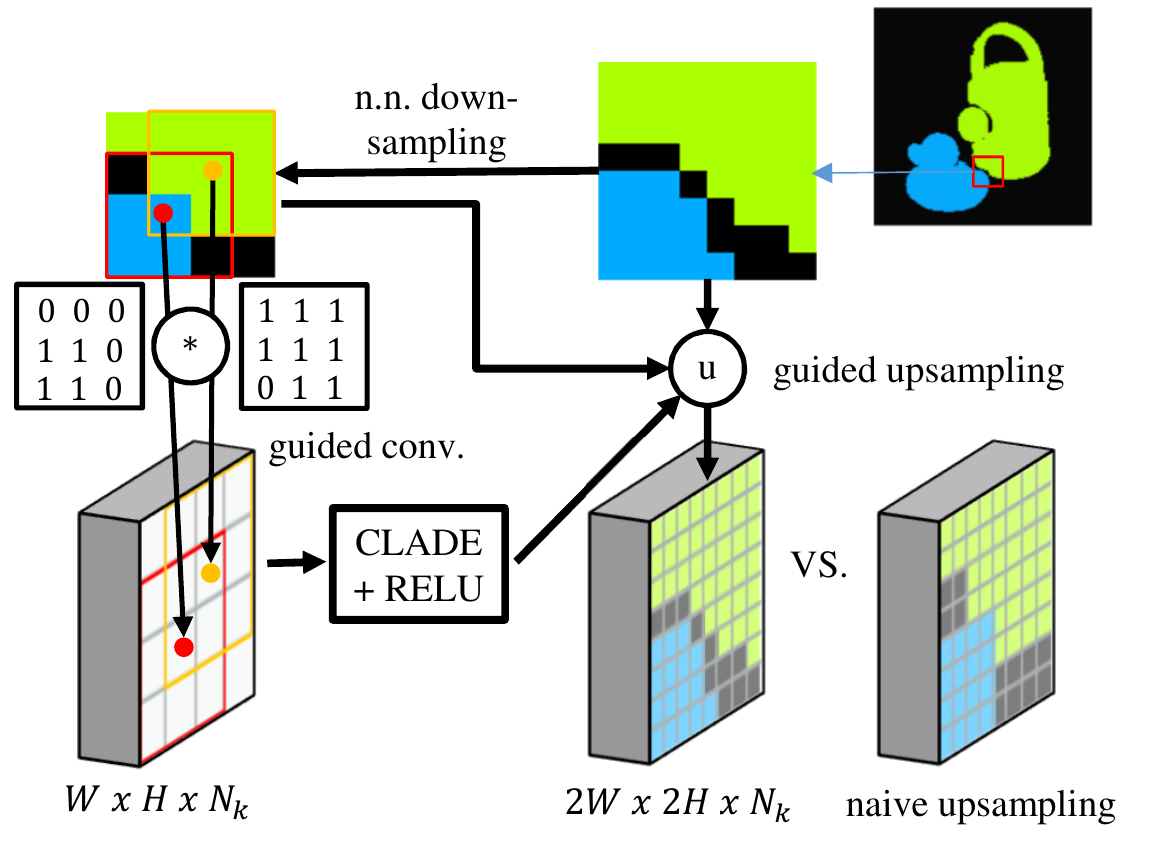}}
 \caption{a) The segmentation is used to select modulation parameters from the weight matrices $\Gamma$ and $B$. b) Guided operations improve the alignment between features and mask.}
 \end{figure}

To reach semantic-awareness, we follow the idea of class-adaptive (de)nor\-ma\-lisation (CLADE) \cite{Tan2020} and make the modulation parameters $\gamma$ and $\beta$ functions of the input segmentation. Thereby, a set $\Gamma=(\gamma^1_k, \dots \gamma^{N_c}_k)$ of scale, and a set $B=(\beta^1_k, \dots \beta^{N_c}_k)$ of shift parameters is learned. The intermediate semantic segmentation is used by Guided Sampling \cite{Tan2020} to convert the sets to dense modulation tensors. Corresponding to the segmentation map, the modulation tensors are filled with the respective parameters, as shown in Fig.~\ref{fig:clade}.

Guided Sampling uses a discrete index to select a specific row from a matrix of de-normalisation parameters. The required $argmax$ operation over the estimated class probabilities sacrifices differentiability, required for end-to-end training. To keep it, we directly use the segmentation input $S = (s_{x,y,l}) \in \mathbb{R}^{H\times W\times N_c}, s \in (0,1)$. The parameter sets are the matrices $\Gamma = (\gamma_{l,k})\in \mathbb{R}^{N_c \times N_k}$ and $B =(\beta_{l,k})\in \mathbb{R}^{N_c \times N_k}$, and the dense modulation tensors are the scalar product over the last dimension of $S$ and the parameter matrices $\Gamma$ and $B$.

\begin{equation}
\bar{\gamma}_{x,y,k} = \sum_{l=0}^{N_c} s_{x,y,l} \gamma_{l,k} ,\quad \bar{\beta}_{x,y,k} = \sum_{l=0}^{N_c} s_{x,y,l} \beta_{l,k} 
\end{equation}

Since this operation should simulate a discrete selection, the predicted label $s_{x,y,l}$ must be either close to 1 or 0. The raw intermediate segmentation $\hat{S} = (\hat{s}_{x,y,l})$ is normalised with a softmax function scaled with a temperature parameter $\tau$ to push all but one value close to 0.
\begin{equation}
s_{x,y,l} = softmax(\tau \hat{s}_{x,y,l})
\end{equation}

\subsection{Semantically Guided Decoder}
\label{sec:guided}
Our goal is to synthesise keypoint vector fields, which are coherent and locally limited for each object. Based on the object a pixel belongs to, the vector should point to a particular location on that object. However, of course, occluding objects should not influence each other decreasing the accuracy in overlapping regions. We use pixel-wise object-specific weights to discriminate the objects from each other, and identify two challenges for the decoder.
\begin{enumerate}  
\item Due to the spatial invariance of convolution, at object boundaries the decoder has no information which parameters were used to normalise a position in the previous block.
\item A nearest neighbour (NN) upsampling after CLADE does not result in a feature map that perfectly matches the higher resolution semantic map in the next block.
\end{enumerate}

Inspired by Dundar \etal \cite{Dundar2020}, we add segmentation-awareness to convolution and upsampling (Fig.~\ref{fig:guided}). The \textbf{object-aware convolution} ensures that the result of a convolution for an object only depends on feature values belonging to it. A mask $M$ filters weights $W$ for every image patch. The mask defines which locations contribute to the feature values and depends on the semantic segmentation $S$. To preserve differentiability, we avoid $argmax$ and do not use a hard binary mask. 

\begin{equation}
\label{equ:guided}
m_{x,y}(i,j) = \begin{cases}
\bar{s}_{x+i,y+j} & \parbox[t]{4.0cm}{ \text{if} \quad $s_{x+i,y+j,l}  = \bar{s}_{x+i,y+j}$  \text{with} \quad $\{ l | s_{x,y,l} = \bar{s}_{x,y} \}$ }\\
0& \text{otherwise}
\end{cases}
\end{equation}
In Equ.~\ref{equ:guided}, $\bar{s}$ is the maximum value along the class dimension of $S$. The indices $x,y$ correspond to the filter location in the image, $i,j$ to the position in the filter.
We apply the object-aware convolution with a $3\times3$ filter at every location by  element-wise multiplication of the input features $X$ with mask $M$ before filtering, followed by normalisation.  
\begin{equation}
x' = W^T (X \odot M ) \frac{9}{sum(M)}
\end{equation}

The \textbf{object-aware upsampling} layer enlarges the feature map without losing the alignment with $S$. Initially, $S$ is NN down-sampled $n$ times ($n=3$) to be used with features in different resolutions. The segmentation in the next higher and current resolution, $S^u$ and  $S^d$, guide the upsampling from low to high resolution features, $F^d$ and $F^u$ \cite{Dundar2020}. This preserves the spatial layout in the feature map after the object-aware convolution and CLADE. We do not apply hole-filling, and select the first feature in the $2\times2$ windows for unknown locations.

\subsection{Differentiable Keypoint Regression}
\label{sec:keypoint}
The local processing of the latent map adds semantic-awareness to process different regions independently. We use this property to avoid the non-differentiable RANSAC estimation, commonly used to get 2D points from vector fields \cite{Peng2019}. 
While exact intersection points can only be calculated for line pairs, a least squares solution can approximate them from multiple vectors \cite{Traa2013}. Each vector inside a semantic region adds one equation to a corresponding linear system. Solving with the Moore-Penrose pseudoinverse guarantees a unique solution. An estimated per-pixel confidence captures the probability that a vector points in the right direction. Weighting equations with the estimated confidence reduces the susceptibility to noise \cite{Traa2013}. An additional loss minimises the euclidian distance between the calculated 2D coordinate and ground truth. By this, the network learns in which regions the most accurate vectors are predicted and boosts their weighting in the calculation. We observe a focus on nearby regions as well as the object contour (Fig.~\ref{fix:example}). A custom layer calculates the intersection points in parallel on the GPU, so that the network outputs the 2D locations directly. The confidence maps increase the number of outputs by the number of keypoints per object.

During inference, we cluster the semantic maps in connected components and solve the system for the largest component per class. By this, potential misdetections in the semantic map can be filtered out. This works very well, if only one instance of each object is visible per image. For an extension to instance segmentation the semantic segmentation encoder would have to be replaced by an instance segmentation decoder, e.g.~\cite{Cheng2020}. Since the components from Section \ref{sec:clade} and \ref{sec:guided} can also be applied to instance masks, all their advantages remain.

\subsection{Merged Network Outputs}
The object-specific weights in the CLADE layers serve as a key in the decoder to decipher the encoded weights in a characteristic manner. The local processing by the object-aware operations minimises performance degradation and crosstalk between objects. The loss function is applied on the fused output directly. The number of channels of the keypoint decoder is constant and only one channel is added per object in the segmentation decoder. For each keypoint, we compute a confidence map and a 2D vector field. This results in  $3m+n+1$ output channels for $n$ objects with $m$ keypoints, e.g.~$41$ for $13$ objects (and 9 keypoints), compared to $248$ outputs for PVNet \cite{Peng2019} or $3342$ for EPOS \cite{Hodan2020}. The constant number of outputs reduces the GPU memory footprint and quickens inference and training.

\section{Implementation Details}
\label{sec:details}
\paragraph{Architecture} 
In CASAPose, a shared ResNet-18 \cite{He2016,Peng2019} provides features for two decoders. The first resembles \cite{Peng2019} and predicts a semantic mask by multiple blocks of consecutive skip connections, convolutions, batch normalisation (BN), leaky ReLU, and upsampling. The keypoint decoder is similar, but replaces BN with CLADE, the regular convolution with an object-aware convolution, and the blind upsampling with object-aware upsampling. Each block takes a scaled semantic mask as additional input to guide the replaced layers. We observed improved convergence if the ground truth segmentation instead of the output of the keypoint decoder is used for supervision. Both decoders can thereby calculate their result in parallel, which results in faster training. The semantic mask, the vector field, and the confidence maps are passed to a keypoint regressing layer (Section \ref{sec:keypoint}). It outputs 9 2D keypoints for each object. Their 3D locations were initially calculated using  the farthest point sampling (FPS) algorithm. The pose is estimated with OpenCV's EPnP \cite{Lepetit09} in an RANSAC scheme. Due to its lightweight backbone our network is small and has only $\approx14.8$ million weights. The CLADE layers increase the total number of weights by only 1024 per object.  
\vspace{-3pt}
\paragraph{Training Strategy}
During training, all but one of the scenes from the Bop Challenge 2020 \cite{Hodan2021} synthetic LINEMOD dataset are used. The images are rendered nearly photo realistically with physically-based rendering (\textit{pbr}). Object and scene parameters, e.g.~object, camera and illumination position, background texture, and material are randomised. The objects are randomly placed on a flat surface with mutual occlusions. We narrow the domain gap by strong augmentation (contrast, colour, blur, noise). We follow Thalhammer \etal \cite{Thalhammer2021} but vary the gain for sigmoid contrast \cite{Jung2021} to be $\mathcal{U}(5, 10)$, since the listed values corrupt the image too much. Each network is trained for 100 epochs using the Adam optimiser \cite{Kingma2014} and a batch size of 18 on two NVIDIA A100 GPUs (4 is the maximum for a single Nvidia RTX 2080Ti). The initial learning rate of $0.001$ is halved after 50, 75 and 90 epochs. We use  \textit{smooth} $\ell1$ loss $\mathcal{L}_{Vec}$ and Differentiable Proxy Voting Loss (DPVL) \cite{Yu2020} $\mathcal{L}_{PV}$ to learn the unit vectors, and Cross Entropy Loss with Softmax $\mathcal{L}_{Seg}$ the learn the segmentation. Additionally, the keypoint loss $\mathcal{L}_{Key}$ is defined as the \textit{smooth} $\ell1$ of the average Euclidean distance between the estimated keypoints and the ground truth keypoints. The overall loss is 
\begin{equation}
\mathcal{L} = \lambda_1 \mathcal{L}_{Seg} + \lambda_2 \mathcal{L}_{Vec} +\lambda_3 \mathcal{L}_{PV} + \lambda_4 \mathcal{L}_{Key}
\end{equation} with $\lambda_1 = 1.0$ , $\lambda_2 = 0.5$, $\lambda_3 = 0.015$ and $\lambda_4 = 0.007$. The $\lambda$ values were determined in empirical preliminary studies using the unseen pbr scene. 
\vspace{-3pt}

\section{Experiments}
\label{sec:eval}
We use test images and objects from the widely used datasets LINEMOD (LM) \cite{Hinterstoisser2012}, Occluded LINEMOD (LM-O) \cite{Brachmann2014}, and HomebrewedDB (HB) \cite{Kaskman2019}. We avoid real camera images in training because for practical applications capturing and annotation of large multi-object datasets is unrealistically costly. Our network estimates the poses of all detected objects in one pass. We report the standard metrics \textbf{ADD} and \textbf{ADD-S} \cite{Xiang2018} for symmetric (glue and eggbox) objects, and the \textbf{2D projection} \cite{Hinterstoisser2012} (\textbf{2DP}) metric. For \textbf{ADD/S}, the average 3D distance (between ground truth (gt) and transformed vertices) must be smaller then 10\% of the object diameter; for \textbf{2DP} the 2D distance (between projected gt and transformed vertices) must be smaller than 5 pixel, for poses to be considered as correct. We always list the \textbf{recall} of correct poses per object. Additional results can be found in the Appendix.

\subsection{Comparison with the State of the Art}
\label{sec:exp_comp}
\paragraph{{Occluded LINEMOD (LM-O) \cite{Brachmann2014}}} 
Table \ref{tab:multi} compares the \textbf{ADD/S} metric on LM-O against other papers which only train on synthetic data or use additional unlabelled images from the test domain \cite{Li2021,Wang2020,Yang2021}. CASAPose achieves 27.8\% better results than the other single-stage multi-object approach PyraPose \cite{Thalhammer2021}. In contrast to their finding, favouring patch based approaches over encoder decoder networks, we show that it is possible to achieve better results with a considerably smaller encoder-decoder network. CASAPose performs 3.8\% better than SD-Pose \cite{Li2021}, a state-of-the-art approach using entirely synthetic training data, but we only train a single network for all objects and omit pre-detection and preprocessing. 
DAKDN \cite{Zhang2021}, the best weakly supervised approach is outperformed by 6.5\%. Like us, the Top-2-6 approaches \cite{Zhang2021,Li2021,Thalhammer2021,Wang2020,Yang2021} all use physically-based rendered (\textit{pbr}) images in training. Compared to approaches using simpler synthetic data, namely CDPN \cite{Zhigang2019} and DPOD \cite{Zakharov2020} our algorithm achieves far superior results. The result for 'eggbox' is weaker than for the other methods. Its symmetry leads to ambiguities of the 2D keypoint projections during training. Future work might consider adding a differentiable renderer and a projection, e.g.~edge-based loss, to explicitly account for symmetry. In addition, 'eggbox' is often heavily occluded, and inclusion of real images (e.g.~semi-supervised) could improve its detection rate.
For EPOS \cite{Hodan2020}, another single-stage approach, only the Average Recall (\textbf{AR}) \cite{Hodan2021} metric is listed with a value of 44.3, whereas our presented results correspond to an \textbf{AR} of 54.2.\footnote{The \textbf{AR} metric is usually used in Bop Challenges \cite{Hodan2021}, in which results different from \cite{Hodan2020} were obtained; see Appendix \ref{sec:experiment_bop} for more details.}. \vspace{3pt} 

\paragraph{LINEMOD (LM) \cite{Hinterstoisser2012}} 
Table \ref{tab:single} compares our result on LM with one network trained for 13 objects to other methods using only synthetic training data. The achieved mean \textbf{ADD/S} of 68.1\% is 1.2\% better than SD-Pose \cite{Li2021} and 7.4\% better than PyraPose \cite{Thalhammer2021}.\vspace{3pt}

\paragraph{HomebrewedDB (HB) \cite{Kaskman2019}} 
The second sequence of HB is a benchmark to check whether a method generalises well to a novel domain \cite{Kaskman2019}. It contains three objects from LM, captured with a different camera in a new environment. Our network trained for LM is used without retraining and the result far surpasses the next best method DAKDN \cite{Zhang2021} by 35\% (Table \ref{tab:homebrewed}). By focusing on the object mask and because 2D-3D correspondences are predicted instead of a full 6D pose, our method shows high invariance to new environments and different capture devices. To verify the latter, another version of the same sequence captured with Kinect (also part of HB) is checked. The camera parameters differ more, but the result is almost as good.

\begin{table}[t!]
\small
\setlength\tabcolsep{2 pt}
\begin{center}
 \begin{tabular}{l|c|c|cccccccc|r } 
\toprule
{Method} & Data & single-st. & Ape  & Can & Cat & Drill & Duck & Eggb. & Glue & Hol.& \textbf{Avg.}   \\ 
\midrule
DPOD\cite{Zakharov2020} & syn. & \checkmark & 2.3 & 4.0 & 1.2 & 10.5 & 7.2 & 4.4 & 12.9 & 7.5 & 6.3 \\
CDPN\cite{Zhigang2019} & syn. & - &  20.0 & 15.1 & 16.4 & 5.0 & 22.2 & 36.1 & 27.9 & 24.0 & 20.8\\
DSC-PoseNet\cite{Yang2021} & pbr+RGB & - &  9.1 & 21.1 & \textbf{26.0} & 33.5 & 12.2 & 39.4 & 37.0 & 20.4 & 24.8 \\
PyraPose\cite{Thalhammer2021}& pbr & \checkmark &  18.5 & 46.4 & 11.7 & 48.2 & 19.4 & 16.7 & 30.7 & 33.0 & 28.1  \\
Self6D\cite{Wang2020} & pbr+RGBD & - &  13.7 & 43.2 & 18.7 & 32.5 & 14.4 & \textbf{57.8} & 52.3 & 22.0 & 32.1 \\
DAKDN\cite{Zhang2021} & pbr+RGB & - & - & - & - & - & - & - & - & - & 33.7\\
SD-Pose\cite{Li2021} & pbr  & - &  21.5 & 56.7 & 17.0 & 44.4 & \textbf{27.6} & 42.8 & 45.2 & 21.6 & 34.6 \\
\midrule
\textbf{CASAPose} & pbr & \checkmark &  \textbf{24.3} & \textbf{59.5} & 15.2 & \textbf{57.5} & 26.0 & 14.7 & \textbf{55.4} & \textbf{34.3} & \textbf{35.9}  \\
\bottomrule
\end{tabular}
\end{center}
\caption{\label{tab:multi} Comparison of ADD/S-Recall with SoTA approaches on LM-O with synthetic only or weakly supervised training using unlabeled real data.}
\end{table}

\begin{table}[t!]
\small
\setlength\tabcolsep{2 pt}
\begin{center}
 \begin{tabular}{lccccccccccccc|r } 
\toprule
{Method} & Ape & Bv. & Cam & Can & Cat & Drill & Duck & Eggb. & Glue & Hol. & Iron & Lamp.& Ph.& \textbf{Avg.}   \\ 
\midrule
AAE \cite{Sundermeyer2019} & 4.2 & 22.9 & 32.9 & 37.0 & 18.7 & 24.8 & 5.9 & 81.0 & 46.2 & 18.2 & 35.1 & 61.2 & 36.3 & 32.6\\  
MHP\cite{Manhardt2019} & 11.9 & 66.2 & 22.4 & 59.8 & 26.9 & 44.6 & 8.3 & 55.7 & 54.6 & 15.5 & 60.8 & - & 34.4 & 38.8\\ 
Self6D-LB\cite{Wang2020} & 37.2 & 66.9 & 17.9 & 50.4 & 33.7 & 47.4 & 18.3 & 64.8 & 59.9 & 5.2 & 68.0 & 35.3 & 36.5 & 40.1\\ 
DPOD\cite{Zakharov2020} & 35.1 & 59.4 & 15.5 & 48.8 & 28.1 & 59.3 & 25.6 & 51.2 & 34.6 & 17.7 & 84.7 & 45.0 & 20.9 & 40.5\\
PyraPose\cite{Thalhammer2021}& 22.8 & 78.6 & 56.2 & 81.9 & 56.2 & 70.2 & 40.4 & 84.4 & 82.4 & \textbf{42.6} & \textbf{86.4} & 62.0 & 59.5 & 63.4\\
SD-Pose\cite{Li2021} & \textbf{54.0} & 76.4 & 50.2 & 81.2 & \textbf{71.0} & 64.2 & \textbf{54.0} & \textbf{93.9} & \textbf{92.6} & 24.0 & 77.0 & 82.6 & 53.7 & 67.3 \\
\midrule
\textbf{CASAPose} & 30.3 & \textbf{94.8} & \textbf{60.0} & \textbf{83.9} & 60.5 & \textbf{89.2} & 37.6 & 71.0 & 80.7 & 30.7 & 84.5 & \textbf{89.9} & \textbf{71.7} & \textbf{68.1}  \\
\bottomrule
\end{tabular}
\end{center}
\caption{\label{tab:single} Comparison of ADD/S-Recall with SoTA approaches on LM with synthetic training.}
\end{table}

\begin{table}[t!]
\small
\setlength\tabcolsep{2 pt}
\begin{center}
 \begin{tabular}{lccccccc|cc } 
\toprule
{Method} & DPOD\cite{Zakharov2020}& PyraP.\cite{Thalhammer2021}  & DSC-P.\cite{Yang2021} & Self6D\cite{Wang2020} & DAKDN\cite{Zhang2021} & ~ & \textbf{Ours}~ & ~& Ours\textsuperscript{\textdaggerdbl} \\ 
\midrule
Avg. & 32.7 & 41.3\textsuperscript{\textdagger} & 44.0 & 59.7\textsuperscript{\textdagger} & 63.8 & ~ &\textbf{86.3}~ & ~& 84.4 \\  
\bottomrule
\end{tabular}
\end{center}
\caption{\label{tab:homebrewed} Comparison of ADD/S-Recall on HB. Methods indicated with (\textsuperscript{\textdagger}) retrain with data from the target domain. We list results for the Primesense and Kinect (\textsuperscript{\textdaggerdbl}) sequence.}
\end{table}

\subsection{Ablation Study}
\label{sec:exp_abl}
Table \ref{tab:ablation} shows the effects of different components of our architecture. We train models for 13 objects and evaluate on LM, LM-O and the unseen \textit{pbr} scene \cite{Hodan2021} to see improvements with and without domain gap. The simplest model (\textit{Base}) uses the merged vector field, but otherwise resembles PVNet \cite{Peng2019}. The model is gradually expanded to include the second decoder with CLADE (\textit{C}) and the semantically guided operations (\textit{C/GCU}). 
For each case, we train a network with and without confidence output, i.e. with Differentiable Keypoint Regression (\textit{DKR}) or RANSAC based voting (\textit{RV}). Averaged over the three datasets, adding \textit{C} and \textit{C/GCU} improves \textbf{ADD/S} by 13.6\% and 17.9\% compared to \textit{Base} for the \textit{RV} networks. This demonstrates that extended network capacity with \textit{C} and the enforced local processing with \textit{GCU} improve the quality of the estimated vector fields. Adding \textit{DKR} further improves \textbf{ADD/S} by 16.1\% for \textit{C} and 17.6\% for \textit{C/GCU}, compared to the respective network with \textit{RV}. The total improvement from Base with \textit{RV} to the final network is 38.7\% for \textbf{ADD/S} and 7.14\% for \textbf{2DP}. Adding \textit{DKR} to \textit{Base} also improved \textbf{ADD/S} by 24.5\%, and \textbf{2DP} by 4.3\% showing that its ability to assign low confidences e.g.~at overlapping regions improves already the simplest architecture. 
We notice that adding \textit{GCU} to a network using \textit{DKR} is especially effective for the datasets with domain gap and even more if occlusion is present (9.3\% and 7.3\%  improvement over \textit{C}+\textit{DKR} compared to 2.7\% improvement for \textit{pbr}). The reason we give for this is that the contour is a cross-domain feature, and access to the silhouette and consequent higher weighting of nearby vectors helps bridging the domain gap. Fig. \ref{fix:example} shows this as well as the sharp separation of the vector fields at object boundaries due to \textit{C/GCU}.

\begin{table}
\setlength\tabcolsep{2.6 pt}
\small
\begin{center}
\begin{tabular}{l c  c  c  c  c  c  c  c c  c  c  c}
{Arch.} &   \multicolumn{6}{c}{\textit{without keypoint regression (RV)}} & \multicolumn{6}{c}{\textit{with keypoint regression (DKR)}} 
\\\cmidrule(lr){2-7}\cmidrule(lr){8-13}
{} &   \multicolumn{2}{c}{LM-O}& \multicolumn{2}{c}{LM} & \multicolumn{2}{c}{pbr} & \multicolumn{2}{c}{LM-O}& \multicolumn{2}{c}{LM} & \multicolumn{2}{c}{pbr} \\      
{} & 2DP & ADD/S & 2DP & ADD/S & 2DP & ADD/S & 2DP & ADD/S & 2DP & ADD/S& 2DP & ADD/S\\
\midrule
Base           & 45.3 & 22.2 & 90.0 & 49.8 & 74.6 & 42.2 & 49.4 & 28.9 & 91.7 & 59.2 & 77.9 & 52.5 \\
C              & 47.3 & 26.6 & \textbf{91.8} & 55.7 & 76.5 & 47.5 & 51.4 & 29.9 & 93.6 & 64.7 & 79.3 & 56.1\\
\textbf{C/GCU} & \textbf{49.2} & \textbf{26.7} & 91.4 & \textbf{59.0} & \textbf{77.3} & \textbf{49.0} & \textbf{51.5} & \textbf{32.7} & \textbf{93.8} & \textbf{68.1} & \textbf{79.6} & \textbf{57.6} \\
\bottomrule
\end{tabular}
\vspace{3mm}
\caption{\label{tab:ablation} Ablation study: Comparison of different network architectures on different datasets.}
\end{center}
\vspace{-5mm}
\end{table}

\subsection{Influence of the Object Number}
\label{sec:exp_obj}
The multi-object capacity of CASAPose with respect to the \textbf{2DP} and \textbf{ADD/S} metric is evaluated in Table \ref{tab:capacity}. Poses are estimated for the 8 LM-O objects on 
LM and LM-O. While the 8-object network on LM is about as good as the 13-object network, it performs 8\% better on LM-O.
Splitting the objects in two groups of four (ape-drill, duck-holepuncher) leads to a further increase of both metrics. The two-network solution increases \textbf{ADD/S} on LM-O to 38.7\%, further extending the distance to the methods in Table \ref{tab:multi}. This increase can be partially explained by the exclusion of symmetrical objects from the first group for which we also observed a stronger enhancement. Future experiments might evaluate the use of a more recent segmentation backbone to further narrow down the observed multi-object gap.

\subsection{Running Time}
\label{sec:exp_perf}
The runtime from image input to final poses for all visible objects in LM-O is 37.3 ms on average with the 8-object model. It splits in 18.8 ms for network inference, 1.7 ms for \textit{DKR}, about 2.9 ms for PnP, and 13.9 ms for finding the largest connected-component (CC) for each class. This results in about 27 frames per second on our test GPU (Nvidia A100). We propose to replace the CC analysis with an instance centre prediction (e.g.~like \cite{Cheng2020}) in future work to add instance awareness to our method and replace the cost intensive processing.

\begin{table}[t]
\setlength\tabcolsep{1.5 pt}
\small
\parbox{.41\linewidth}{
\centering
\begin{tabular}{l c c c c  }

\toprule
{} &  \multicolumn{2}{c}{LM-O} & \multicolumn{2}{c}{LM}\\
\cmidrule(lr){2-3}\cmidrule(lr){4-5}
{}   & 2DP   & ADD/S    & 2DP   & ADD/S \\\midrule
13 Obj. & 51.5 & 32.7 & 96.0 & 60.4   \\
8 Obj.  & 54.7 & 35.9 &  96.9 & 59.2 \\
4 Obj. (2x) & \textbf{58.4}  & \textbf{38.7}  & \textbf{97.3} & \textbf{64.5}\\
\bottomrule
\end{tabular}
\vspace{2mm}
\caption{\label{tab:capacity} Influence of object number per network on ADD/S-Recall. }
}
\hfill
\parbox{.55\linewidth}{
\centering
\includegraphics[width=\linewidth]{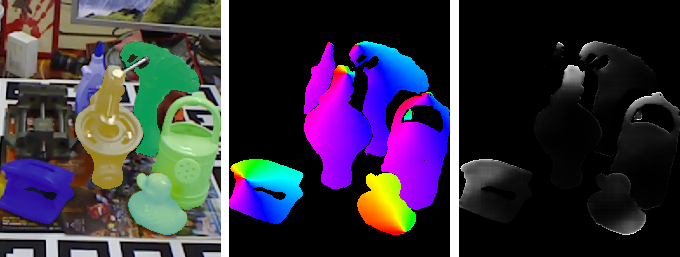}
\vspace{-5mm}
\captionof{figure}{\label{fix:example} Estimated masks, colour coded vector fields, and confidence maps with \textit{C/GCU} (f.l.t.r). }
}
\end{table}

\section{Conclusion}
We showed that class-adaptiveness and semantic awareness improve the performance of a multi-object 6D pose estimator. Local feature processing minimises interference between overlapping regions in an reduced output space. Object-specific parameters selected via CLADE in a second decoder strengthen the prediction accuracy. The locality of the operations allows region-wise predictions, e.g.~of least square weights, where we demonstrated a reduction of domain gap with our Differentiable Keypoint Regression. This also enables the direct addition of additional steps, such as pose refinement, in an end-to-end solution in future work. The presented layers are general enough to be integrated also in other pose estimation architectures.


\section*{Acknowledgements}

This work is supported by the German Federal Ministry for Economic Affairs and Climate Action (BIMKIT, grant no. 01MK21001H). 

%
%
\bibliography{literature}

\newpage
\section*{{\LARGE Appendix}}
\appendix
\noindent We provide visual results of CASAPose in Section \ref{sec:visual}. Section \ref{sec:more_experiments} contains further comparisons with the state of the art as well as two more experiments. Finally, Section \ref{sec:additional_details} lists additional implementation details.

\section{Visual Results}
\label{sec:visual}
\begin{figure}[h!]
\centering     
\subfigure{\label{lmo_fig:a}\includegraphics[width=40mm]{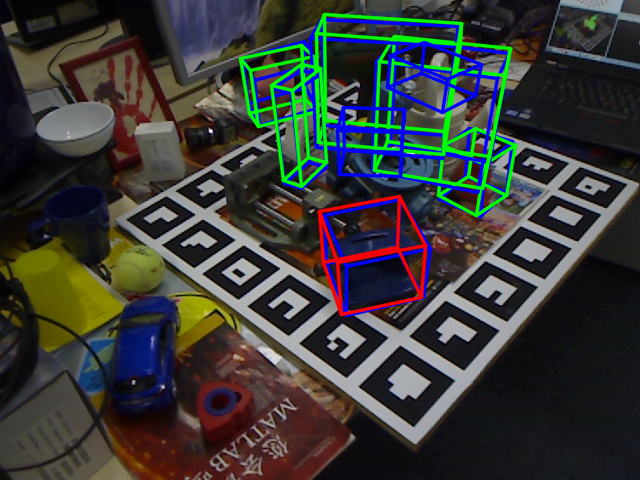}}
\subfigure{\label{lmo_fig:b}\includegraphics[width=40mm]{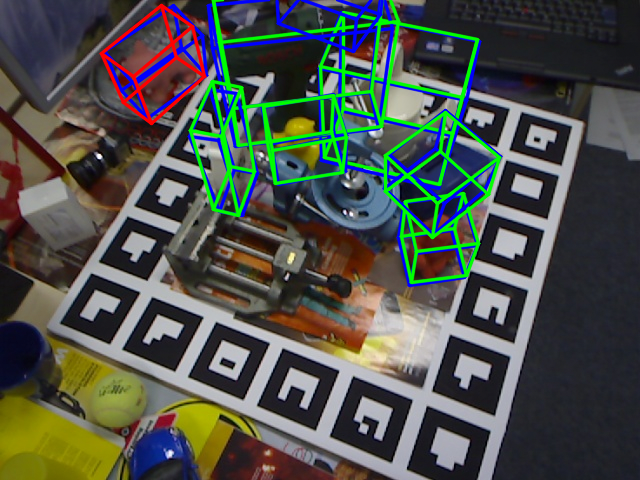}}
\subfigure{\label{lmo_fig:c}\includegraphics[width=40mm]{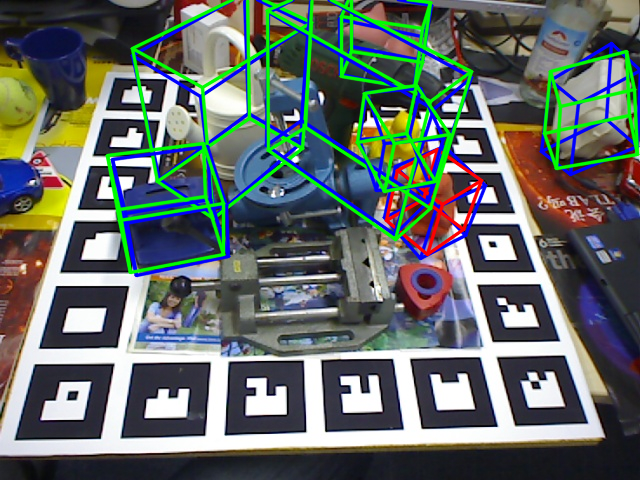}}
\caption{Example results of CASAPose for LM-O, with bounding boxes for correctly estimated poses in green, incorrect poses in red, and ground truth poses in blue (ADD/S metric).}
\label{results_lmo}
\end{figure}

\begin{figure}[h!]
\centering     
\subfigure{\label{pbr_fig:a}\includegraphics[width=40mm]{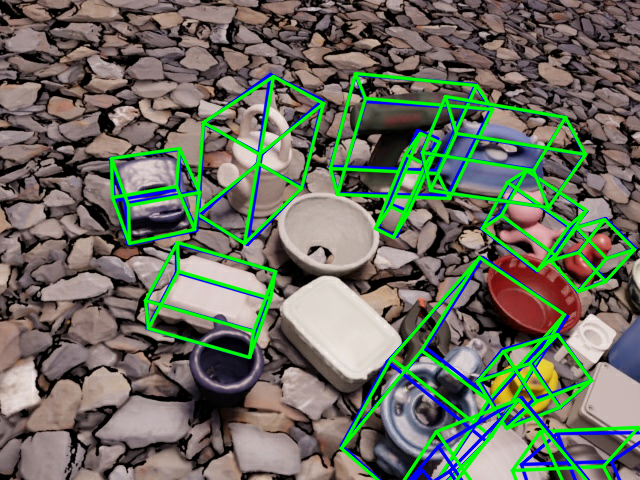}}
\subfigure{\label{pbr_fig:b}\includegraphics[width=40mm]{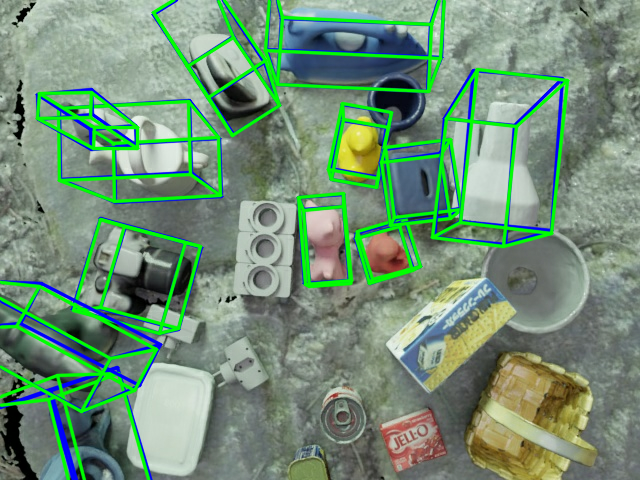}}
\subfigure{\label{pbr_fig:c}\includegraphics[width=40mm]{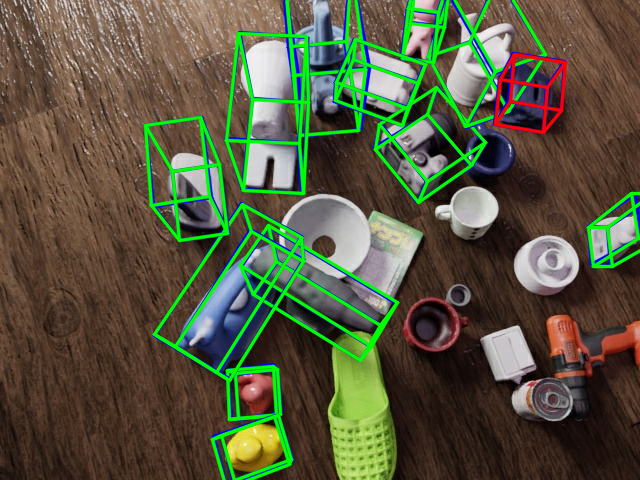}}
\caption{Example results of CASAPose for \textit{pbr} images excluded during training, with bounding boxes for correctly estimated poses in green, incorrect poses in red, and ground truth poses in blue (ADD/S metric).}
\label{results_pbr}
\end{figure}

\begin{figure}[h!]
\centering    
\subfigure{\label{hb_fig:a}\includegraphics[height=40mm]{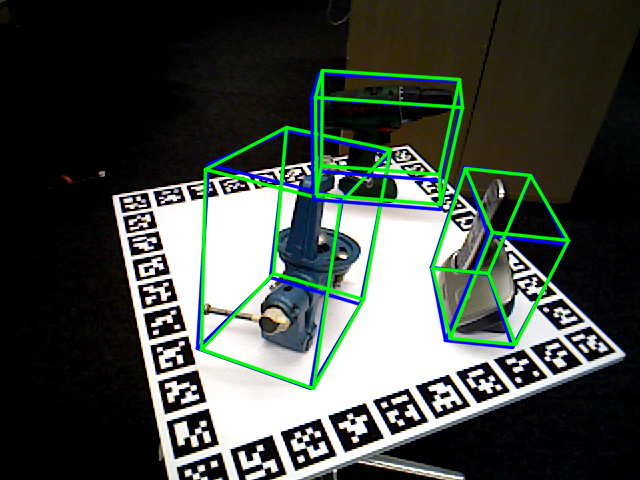}}
\subfigure{\label{hb_fig:b}\includegraphics[height=40mm]{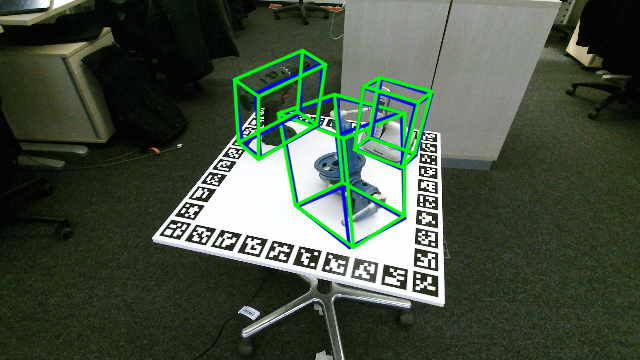}}
\caption{Example results of CASAPose for HB, with bounding boxes for correctly estimated poses in green and ground truth poses in blue (ADD/S metric). The left image is captured with the PrimeSense Carmine camera; the right image is captured with Microsoft Kinect 2.}
\label{results_hb}
\end{figure}

\quad Fig.~\ref{results_lmo} shows estimated poses for three example images from Occluded LINEMOD (LM-O) \cite{Brachmann2014} using CASAPose trained for 8 objects. Similarly, Fig.~\ref {results_pbr} gives an impression of example results using a model trained for 13 objects on our \textit{pbr} test scene \cite{Hodan2021}.
Example results of CASAPose for HomebrewedDB (HB) \cite{Kaskman2019} are shown in Fig.~\ref {results_hb}.

\subsection{Effect of Guided Operations} 
\begin{figure}[h!]
\centering     
\subfigure[input image and estimated mask]{\label{clade_fig:a}\includegraphics[width=.32\linewidth]{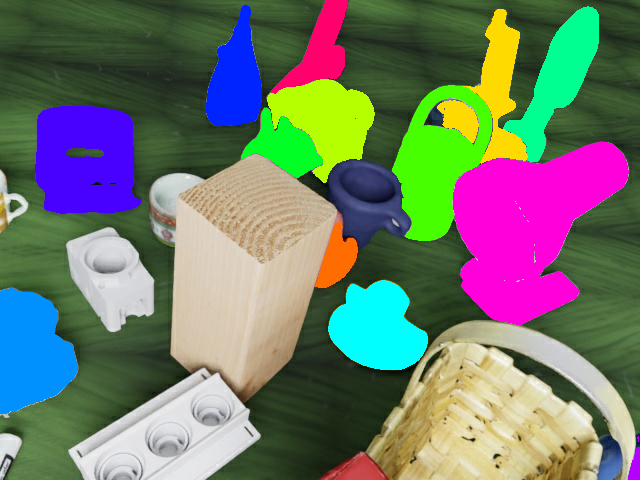}} 
\subfigure[estimated vector field (Base)]{\label{clade_fig:b}\includegraphics[width=.32\linewidth]{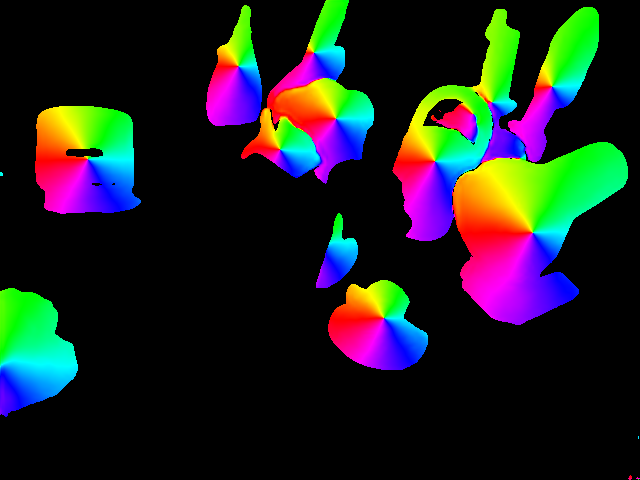}}
\subfigure[estimated vector field (C/GCU)]{\label{clade_fig:c}\includegraphics[width=.32\linewidth]{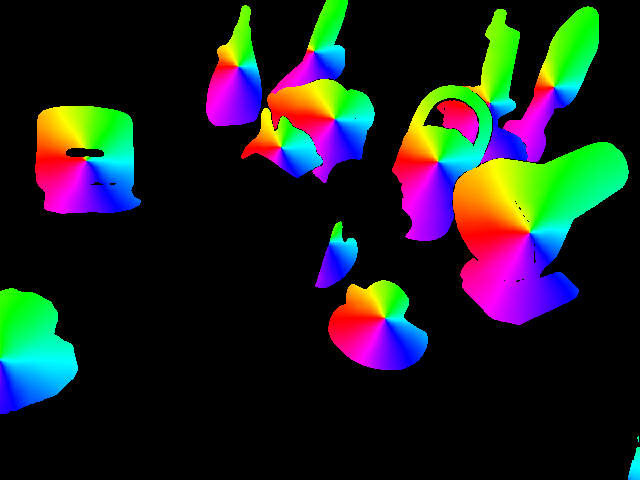}} \\
\subfigure[detail comparison 1]{\label{clade_fig:d}\includegraphics[width=.35\linewidth]{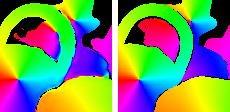}}
\subfigure[detail comparison 2]{\label{clade_fig:e}\includegraphics[width=.35\linewidth]{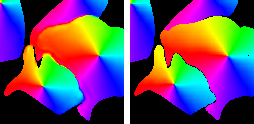}}
\caption{Visual comparison of the estimated vector fields for a network with (\textit{C/GCU}) and without (\textit{Base}) the semantically guided operations. In the detail comparisons, \textit{Base} is on the left, while \textit{C/GCU} is on the right.}
\label{improvements_clade}
\end{figure}

\quad The ablation study discovered that semantic guidance improves the estimated vector fields and the accuracy of pose estimation. The effect is best seen in direct visual comparison. Fig.~\ref{improvements_clade} shows the enhancement exemplified for an image from the \textit{pbr} dataset using colour coded vector fields for visualisation. It shows the vector fields for the first keypoint, which is always located in the centre of each object.
Comparing the output of a model without the guided decoder in Fig.~\ref{clade_fig:b} with the output of a model with object-aware convolutions and object-aware upsampling in Fig.~\ref{clade_fig:c}, there is a clear improvement in vector fields, especially in regions where objects overlap. In fact, we have observed that a network without semantic guidance is not even able to produce perfectly separated vector fields when it heavily overfits only a few images. Using CLADE alone without semantic guidance already improves the quality of the vector fields per object due to the object-specific parameters (see Table \ref{tab:ablation} of the main paper), but a clear separation as in Fig.~\ref{clade_fig:c} can only be achieved in combination. 

\subsection{Characteristics of the Learned Confidence Maps} 
\begin{figure}
  \centering
  \subfigure[Image and estimated mask]{\label{kp:a}\includegraphics[width=.49\linewidth]{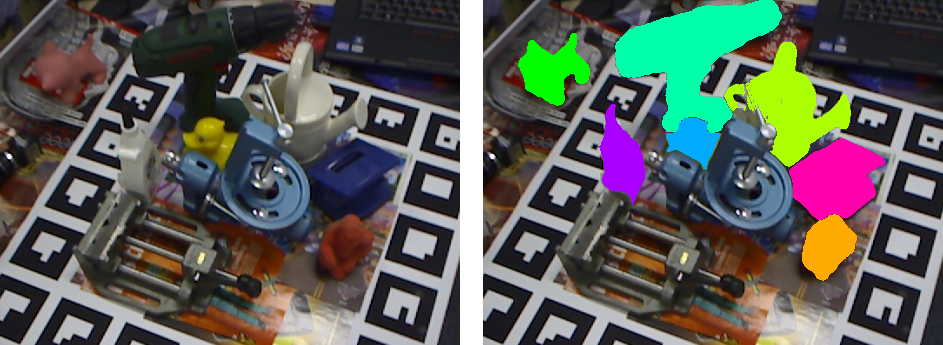}}
  \subfigure[Keypoint \#1]{\label{kp:b}\includegraphics[width=.49\linewidth]{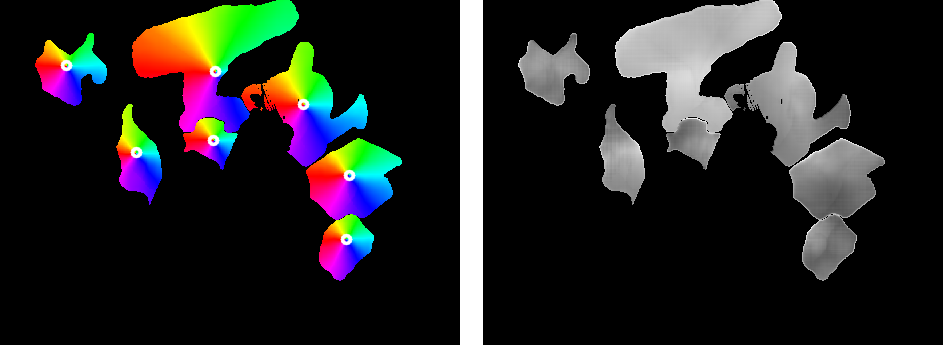}}
    \subfigure[Keypoint \#4]{\label{kp:c}\includegraphics[width=.49\linewidth]{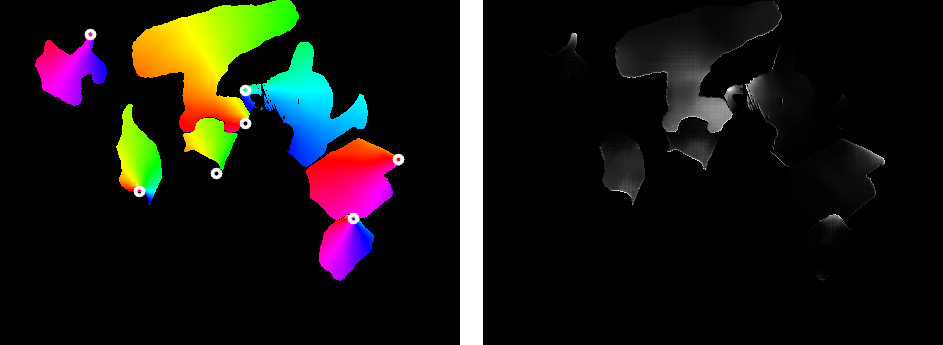}}
    \subfigure[Keypoint \#5]{\label{kp:d}\includegraphics[width=.49\linewidth]{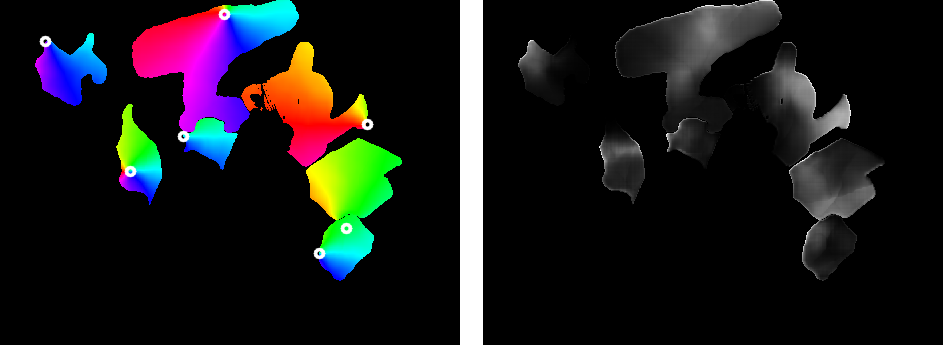}}
  \caption{Estimated vector fields and confidence maps for three out of nine keypoints.}
\label{keypoint_locations}
\end{figure}

Fig.~\ref{keypoint_locations} shows the estimated vector fields and confidence maps for an image from LM-O using the 8-object model. The estimated 2D locations are highlighted by a white circle. The confidence values are normalised inside each semantic mask for clearer presentation. For the first keypoint (Fig.~\ref{kp:b}), which is always in the centre of the object, the confidence is relatively constant in each mask, indicating that it is easy for the network to predict this point with high accuracy. In Fig.~\ref{kp:c} and \ref{kp:d}, it can be seen that the regions where the network predicts high confidence are often spatially close to the actual keypoint location. For example, for the tip of the tail of 'cat' in Fig.~\ref{kp:c}, it is logical that the best prediction of the location can be made nearby.  Moreover, for example, 'ape' in Fig.~\ref{kp:d} shows that the model predicts high reliability and thus computes the 2D position of a keypoint mainly from pixels near the object silhouette. Especially for non-textured objects, the silhouette provides important information about the orientation of the object. It seems appropriate that the vectors near the contour can be estimated with higher precision.

\section{Additional Experiments}
\label{sec:more_experiments}
\subsection{BOP Challenge Evaluation}
\label{sec:experiment_bop}
In the BOP Challenge 2020 \cite{Hodan2021} multiple approaches submitted results for several pose estimation datasets, including LM-O with synthetic training. The results presented are in most cases significantly improved by subsequently inserted changes compared with the results from the original publications. We also evaluated our results against the BOP benchmark. It calculates an accuracy called Average Recall (\textbf{AR}), which is the average of the results for three pose error functions, Maximum Symmetry-Aware Projection Distance (MSPD), Maximum Symmetry-Aware Surface Distance (MSSD), and Visible Surface Discrepancy (VSD). Further details can be found in \cite{Hodan2021}. 

Table \ref{tab_bop} lists the results of our procedure with this metric. CASAPose\textsubscript{8} is our final result for the 8-object case from the main paper. EPOS \cite{Hodan2020, Hodan2021}, the only other single-stage multi-object method achieved an \textbf{AR} of 54.7, slightly higher than CASAPose. CASAPose\textsubscript{8*} trained with minimally different hyper parameters (increasing $\lambda_4$ from $0.007$ to $0.01$), again achieves a slightly superior result, showing that both methods are similarly accurate. Still, our method is multiple times faster (468ms vs. 37 ms) \footnote{The difference is so significant, that it can also not be explained by our faster evaluation GPU.}. CDPNv2 \cite{Zhigang2019, Hodan2021}, the best method using only rgb images and no additional refinement (EPOS would be the second best method on LM-O with these properties), reaches an \textbf{AR} of $62.4$, but trains a single network for every object. They make numerous extensions to their original approach (adding more complicated domain randomisation and a more powerful backbone) that would potentially improve our method as well, but are out of scope of this paper. The innovations of our paper to convert a multi-stage (one network per object and bounding box detector) approach into a single-stage (one network for all objects without need for a bounding box detector) approach could be applied analogously to their method.

\begin{table}[h!]
\setlength\tabcolsep{2.6 pt}
\small
\begin{center}
\begin{tabular}{l c  c  c  c  c }
\toprule
{Arch.} &  DNN  & $AR$ & $AR_{MSPD}$ & $AR_{MSSD}$ & $AR_{VSD}$ \\
\midrule
EPOS \cite{Hodan2020, Hodan2021} & 1/set & 54.7 & 75.0 & 50.1 & 38.9 \\ \midrule
CASAPose\textsubscript{8} & 1/set & 54.2 & 74.3 & 49.4 & 39.0 \\
CASAPose\textsubscript{8*} & 1/set&  55.4 & 75.3 & 50.8 & 40.2 \\  \midrule  \midrule
CASAPose\textsubscript{2x4} & 2/set &  57.4 & 77.1 & 52.9 & 42.1\\
\bottomrule
\end{tabular}
\vspace{2mm}
\caption{\label{tab_bop} Comparison of different variants of our method using the BOP benchmark on LM-O with EPOS \cite{Hodan2020} in BOP configuration \cite{Bop2020, Hodan2021}. CASAPose\textsubscript{2x4} is the \textit{'4 Obj. 2x'} result from Table \ref{tab:capacity} of the main paper using 2 notworks, each for 4 objects.} 
\end{center}
\end{table}

\subsection{Ablation Study: Guided Decoder}
We tested different versions of the semantically guided decoder for the 13-object configuration trained with \textit{DKR} (Table \ref{tab:ablation_guided}). The first variant \textit{C/GU} uses only guided upsampling and no guided convolution. Compared with CLADE (\textit{C}) alone, this does not bring an improvement, since the increased accuracy during upsampling is cancelled out by the following regular convolution, which does not take the masks into account. Adding guided convolutions in the first 3 or 4 of 5 decoder blocks (\textit{C/GCU3}, \textit{C/GCU4}) improves the average \textbf{2DP} and the average \textbf{ADD/S}. Between \textit{C/GCU3} and \textit{C/GCU4} no clear difference is visible. Comparing the final model \textit{C/GCU5} (with guided convolutions in all decoder blocks) with \textit{C/GCU4}, the \textbf{2DP} decreases by 0.6\% on average, while an increase of 4.1\% of \textbf{ADD/S} outweighs this. This makes this architecture the best among the tested ones.

\subsection{Influence of Keypoint Regression}
Table \ref{tab:pose_accuracy} compares different variants of the calculation of 2D keypoint positions. $LS_{1stComp.}$ is the variant used in our final model. It applies \textit{DKR} on the largest connected component of each object class and clearly outperforms RANSAC voting ($PV_{RANSAC}$) \cite{Peng2019} used with the same trained model. Interestingly, if a network learns to estimate confidence maps with \textit{DKR} during training, also the results of the RANSAC voting improve ($PV_{RANSAC}$ compared with $PV_{RANSAC*}$). This suggests that least squares optimisation over all vectors in a region during training also improves the global accuracy of the estimated vectors. 
Applying \textit{DKR} on a complete mask without connected component filtering ($LS_{All}$) deteriorates the performance, indicating that potential clutter in the estimated semantic masks should be removed before calculating the 2D positions. We tested \textit{DKR} on the second largest connected component $LS_{2ndComp.}$ and see that it nearly never leads to a correct pose. So, at least for the case where only one object per class is visible, using only the largest connected components is very suitable. A proposal for adaptation to multi-instance scenarios is given in the main paper. 

\begin{table}
\small
\parbox{.5\linewidth}{
\centering
\begin{tabular}{l c c c c}\toprule
{} &   \multicolumn{2}{c}{LM-O}& \multicolumn{2}{c}{LM}
\\\cmidrule(lr){2-3}\cmidrule(lr){4-5}
{} &  2DP & ADD/S&  2DP & ADD/S\\\midrule
C   &  51.4 & 29.9  &  93.6 &  64.7  \\
C/GU   &  50.7 & 30.0  &  93.9 &  62.9  \\
C/GCU3 &  52.0 & 32.1 &  93.7 &  65.5 \\
C/GCU4 &  \textbf{52.7} & 31.9 & 93.5  &  64.9  \\
\textbf{C/GCU5} &  51.5 & \textbf{32.7} &  \textbf{93.8} &  \textbf{68.1}   \\\bottomrule
\end{tabular}
\vspace{2mm}
\caption{\label{tab:ablation_guided} Comparison of different versions of the semantically guided decoder using the 13-object model with \textit{DKR}.}
}
\hfill
\parbox{.45\linewidth}{
\centering
\begin{tabular}{l c c }\toprule
{} & 2DP  & ADD/S \\
\toprule
$PV_{RANSAC*}$  & 49.2  & 26.7\\
$PV_{RANSAC}$  & 50.4  & 30.8\\\midrule
$LS_{All}$  & 45.3  & 29.7\\
$LS_{2nd Comp.}$ & 7e-3  & 2e-3\\
$LS_{1stComp.}$ & \textbf{51.5}  & \textbf{32.7}\\
\bottomrule
\end{tabular}
\vspace{2mm}
\caption{\label{tab:pose_accuracy}Comparison of different variants of 2D keypoint calculation using the 13-object model (\textit{C/GCU5}) on LM-O.}
}
\end{table}

\section{Additional Details}
\label{sec:additional_details}
\subsection{Differentiable Keypoint Regression}
The Differentiable Keypoint Regression uses a weighted Least Squares intersection of lines  calculation, incorporating confidence scores as weights, as it is described e.g.~in \cite{Traa2013}. One system is constructed per keypoint per object. All systems are solved in parallel using \texttt{tf.linalg.pinv} to calculate the Moore-Penrose pseudo-inverse. \textit{DKR} uses the \textit{softplus} function to translate from the network output to the weights of the Least Squares calculations. It is a smooth approximation of the \textit{ReLU} function that constraints the output of the network to be non negative. Compared with \textit{sigmoid}, it allows to predict weights greater than 1. During training, we add a regularisation term to avoid drift of the Least Squares weights for \textit{DKR} towards zero or infinity. The mean value in the foreground regions of each output map is $\ell1$ regularised to be close to a constant value of 0.7.

\subsection{Hyperparameter Choices}
The losses $\mathcal{L}_{Seg}$, $\mathcal{L}_{Vec}$, $\mathcal{L}_{PV}$, and $\mathcal{L}_{Key}$ are weighted with the factors $\lambda_{1-4}$. Previous work weighted $\mathcal{L}_{Seg}$ and $\mathcal{L}_{Vec}$ equally ($\lambda_1=\lambda_2=1.0$) \cite{Peng2019}, or additionally added $\mathcal{L}_{PV}$ as a regularizer with much smaller weight \cite{Yu2020}. In tests without \textit{DKR} and $\mathcal{L}_{Key}$, we determined $0.015$ as a suitable choice for $\lambda_3$. This is larger than the recommendation of \cite{Yu2020}, but leads to stable convergence in our case. A reduction of the influence of $\mathcal{L}_{Vec}$ ($\lambda_2=0.5$) after including $\mathcal{L}_{Key}$ preserves the balance between segmentation ($\lambda_1$) and vector field prediction ($\lambda_{2-4}$). In summary, the weights were $\lambda_1 = 1.0$ , $\lambda_2 = 0.5$, $\lambda_3 = 0.015$ and $\lambda_4 = 0.007$. 

As reported in Section \ref{sec:experiment_bop}, we also trained a full model using $\lambda_4=0.01$ and observed a slight accuracy increase for LM-O (8 objects). However the 13-object model (LM) did not converge as good in this setting.

For models with two decoders (\textit{C} and \textit{C/GCU}), the calculation of $\mathcal{L}_{Vec}$ and $\mathcal{L}_{PV}$ evaluates only those locations where the estimated segmentation matches the true segmentation, which in our experience slightly improves the training result.

\subsection{Further Details}
\begin{itemize}
\item The Farthest Point Sampling (FPS) algorithm is used to calculate the 3D locations of the keypoints \cite{Peng2019}. The keypoint set is initialised by adding the object centre. The 3D models of HB originate from a different 3D scan and have their origin in a different location than the models of LM. We aligned them with the Iterative Closest Point (ICP) algorithm and calculate a fixed compensation transformation for each model. It is applied to the 3D keypoints to make a comparison with HB's ground truth.
\item Pose estimation uses OpenCV's \texttt{cv::solvePnPRansac} with EPnP \cite{Lepetit09} followed by a call of \texttt{cv::solvePnP} with \texttt{SOLVEPNP\_ITERATIVE} and the previous pose as \texttt{ExtrinsicGuess}.
\item During training, we use scenes $0$-$48$ from the synthetic \textit{pbr} LINEMOD images from \cite{Hodan2021} resulting in 49000 training images. Scene $49$ (1000 images) is kept for testing and is used in the ablation study.
\item The experiments were conducted using Tensorflow 2.9 and use ADAM optimiser \cite{Kingma2014}. Our custom layers use Tensorflow's \texttt{\@tf.function(jit\_compile=True)} for acceleration. 
\end{itemize}

\end{document}